\documentclass[letterpaper, 10 pt, journal, twoside]{IEEEtran} 

\IEEEoverridecommandlockouts                              % This command is only needed if 
\usepackage{amsmath}
\usepackage{amssymb}
\usepackage{booktabs} % for table aesthetics
\usepackage{xcolor}
\usepackage{graphicx}
\usepackage{multirow} 
\usepackage{array}    
\usepackage{makecell} 
\usepackage{float}
\usepackage{caption}
\usepackage{textcomp}
\usepackage[colorlinks=true,linkcolor=black,citecolor=black,urlcolor=blue]{hyperref}
\usepackage{xcolor}
\usepackage[normalem]{ulem} 

\newcommand{\bs}{\boldsymbol}
\newcommand{\mb}{\mathbf}
\newcommand{\Fig}[1]{Fig.~\ref{#1}}

\author{Ziken Huang, Xinze Niu, Bowen Chai, Renbiao Jin, Danping Zou$^{*}$ % <-this % stops a space
\thanks{Manuscript received: July, 22, 2025; Revised October, 21, 2025; Accepted November, 19, 2025.
This paper was recommended for publication by Editor G. Loianno upon evaluation of the Associate Editor and Reviewers' comments. %Use only for final RAL version
This work was supported
by the National Key R\&D Program of China under Grant 2022YFB3903802 and the National
Science Foundation of China under Grant 62073214. (\textit{Corresponding author: Danping Zou.)}}
\thanks{All authors are with School of Automation and Intelligent Sensing, Shanghai Jiao Tong University, Shanghai, 200240, China. (e-mail: \{huangzk168, dpzou\}@sjtu.edu.cn)}%
\thanks{Digital Object Identifier (DOI): see top of this page.}
}

\title{Swooper: Learning High-Speed Aerial Grasping \\ with a Simple Gripper}

\let\oldtwocolumn\twocolumn
\renewcommand\twocolumn[1][]{%
  \oldtwocolumn[{#1%
    \begin{center}
        \vspace{-0.7cm} 
        \includegraphics[width=1.0\linewidth,trim={0.5cm 6cm 0.5cm 5cm},clip]{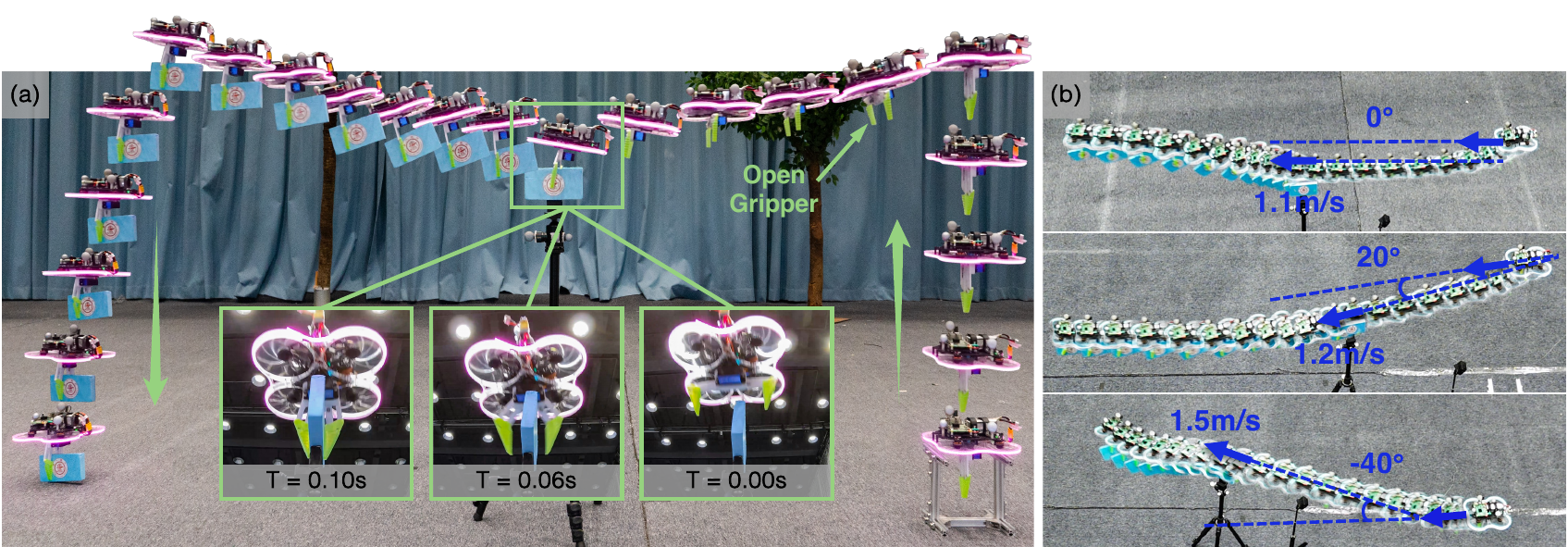}
        \captionof{figure}{Real-world flight snapshots with our deep reinforcement learning policy. (a) Our policy exhibits active manipulation behavior: the agent actively opens the gripper in preparation for grasping and then determines the appropriate timing to close it swiftly (within 0.1s) to enable a smooth and rapid swooping motion for high-speed aerial grasping. (b) Our policy can robustly grasp objects placed at different positions and yaw orientations with grasping speeds of up to 1.5 m/s.}
        \label{fig:realflight}
    \end{center}
  }]
}

\begin{document}
 
\maketitle

%%%%%%%%%%%%%%%%%%%%%%%%%%%%%%%%%%%%%%%%%%%%%%%%%%%%%%%%%%%%%%%%%%%%%%%%%%%%%%%%
\begin{abstract}

High-speed aerial grasping presents significant challenges due to the high demands on precise, responsive flight control and coordinated gripper manipulation. 
In this work, we propose \emph{Swooper}, a deep reinforcement learning (DRL) based approach that achieves both precise flight control and active gripper control using a single lightweight neural network policy. Training such a policy directly via DRL is nontrivial due to the complexity of coordinating flight and grasping. To address this, we adopt a two-stage learning strategy: we first pre-train a flight control policy, and then fine-tune it to acquire grasping skills. With the carefully designed reward functions and training framework, the entire training process completes in under 60 minutes on a standard desktop with an Nvidia RTX 3060 GPU.
% To enable sim-to-real transfer
To validate the trained policy in the real world, we develop a lightweight quadrotor grasping platform equipped with a simple off-the-shelf gripper, and deploy the policy in a zero-shot manner on the onboard Raspberry Pi 4B computer, where each inference takes only about 1.0~ms. In 25 real-world trials, our policy achieves an 84\% grasp success rate and grasping speeds of up to 1.5~m/s without any fine-tuning. 
% This matches the robustness and agility of state-of-the-art systems that rely on traditional pipelines and sophisticated grippers.
This matches the robustness and agility of state-of-the-art classical systems with sophisticated grippers, highlighting the capability of DRL for learning a robust control policy that seamlessly integrates high-speed flight and grasping.
The supplementary video is available for more results.

Video: https://zikenhuang.github.io/Swooper/.

\end{abstract}

\vspace{-0.2cm}
\section{Introduction}

\IEEEPARstart{H}{ow} can we efficiently collect samples in challenging environments, such as rubble zones after nuclear explosions or towering polar glaciers? Aerial grasping, achieved by equipping an unmanned aerial vehicle (UAV) with a specialized gripper, appears to be a promising solution. Compared to ground-based platforms such as legged or wheeled robots, aerial platforms provide significant advantages, especially in scenarios requiring rapid three-dimensional navigation or traversing impassable terrains \cite{bauer_open-source_2024}. Research on aerial grasping also lays a solid foundation for embodied aerial physical interaction \cite{aucone_embodied_2025}.

Recently, Deep Reinforcement Learning (DRL) emerges as a promising approach to the aerial grasping task. Unlike traditional approaches, DRL can directly optimize task-level objectives \cite{song_reaching_2023}, enabling a single policy to unify both flight and gripper control. Recent advances in DRL-based autonomous flight have demonstrated impressive capabilities across various challenging tasks, including drone racing \cite{song_reaching_2023, song_autonomous_2021, penicka_learning_2022, kaufmann_champion-level_2023}, \cite{xing2024bootstrapping}, trajectory tracking \cite{xing2024multi}, flight through narrow gaps \cite{wang_agile_2024, wu_whole-body_2024}, and autonomous landing \cite{kooi_inclined_2021, ladosz_autonomous_2024}. %Unfortunately, it has yet to be validated in aerial grasping. 
In parallel, DRL has also led to substantial progress in grasping on ground robots \cite{huang_earl_2023, hu_grasping_2023, orsula_learning_2022}. 
These developments highlight the strong potential of DRL-based methods for aerial grasping. Therefore, it is of great interest to explore the potential of DRL for training a unified policy that integrates flight and grasping.

However, DRL-based aerial grasping still presents several challenges. First, achieving smooth and rapid grasping requires a flight control policy that is both precise and responsive. Second, reinforcement learning suffers from low sample efficiency, and designing effective reward functions is often nontrivial. While curriculum learning can accelerate convergence, designing an appropriate curriculum requires thorough knowledge of both the task and the agent's abilities. Third, aerial grasping inherently couples navigation with manipulation, further amplifying these challenges. As shown in Section~\ref{subsec:comparison_from_scratch}, training an aerial grasping policy directly from scratch is inefficient and difficult to converge. Finally, policies trained in simulation commonly face significant sim-to-real gaps when deployed on physical UAVs due to dynamics mismatch, latency, and sensor noise.
% One intuitive approach is to employ two separate policies: one for navigation and the other for gripper control, similar to \cite{kooi_inclined_2021}. However, this method introduces pauses during policy switching, preventing rapid dynamic grasping. Moreover, such modular approach lacks comprehensive task-level optimization, failing to fully leverage DRL's representational capabilities. 

We propose \emph{Swooper}: a  DRL-based approach for high-speed aerial grasping. The name reflects the system's aggressive grasping behavior, much like an eagle swooping down to capture its prey. Instead of training directly from scratch, Swooper adopts a two-stage learning strategy by first pre-training a flight control policy and then fine-tuning it to acquire grasping skills. Thanks to the carefully designed reward functions and training framework, the entire training process takes less than 60 minutes on a standard desktop with an Nvidia RTX 3060 GPU, demonstrating high sample efficiency. Consequently, a unified policy is obtained to achieve both precise flight control and active manipulation. In terms of active manipulation, the policy is able to determine the appropriate timing for both opening and closing the gripper to enable seamless grasping during flight, as shown in \Fig{fig:realflight}~(a).
%In terms of active manipulation, the policy is aware of opening the gripper in preparation for grasping and determining the appropriate timing to close it near the target object for seamless grasping.
%This two-stage training strategy not only accelerates convergence but also enhances flexibility, allowing the training of flight control policy tailored to specific performance requirements.

To validate the trained policy in the real world, we develop a lightweight and agile quadrotor grasping platform equipped with a two-finger gripper, and deploy the policy on the onboard computer. Compared with customized soft grippers \cite{fishman_dynamic_2021, Ubellacker2024}, 
our gripper is an off-the-shelf component, which is simple, lightweight, and low-cost.
Although it exhibits lower tolerance to flight control errors, making high-speed aerial grasping more challenging, the trained policy has achieved sim-to-real transfer on our platform without any fine-tuning.
To the best of our knowledge, this work is the first demonstration of high-speed aerial grasping on a physical quadrotor platform using a single policy trained by deep reinforcement learning. 
Our work serves as a stepping
stone towards end-to-end vision-based aerial manipulation.
The main contributions of this work are summarized as follows:

\begin{enumerate}
\item We propose a two-stage DRL-based approach for high-speed aerial grasping, which overcomes the difficulty in training the policy from scratch. 
Our approach unifies precise flight control and active manipulation into a single, lightweight policy.
%first pre-trains a flight control policy and then fine-tunes it for grasping operations.  
%Notably, our approach trains a single, highly lightweight policy to achieve both precise flight control and active manipulation. 

\item We customize a lightweight and agile aerial grasping platform equipped with a simple off-the-shelf gripper. Our policy achieves zero-shot sim-to-real transfer on this platform with an 84\% grasp success rate and grasping speeds of up to 1.5 m/s, 
matching the robustness and agility of state-of-the-art classical systems.

\item We conduct comprehensive simulation and real-world experiments to validate the effectiveness of Swooper.
% We customize a lightweight, low-cost, and agile aerial grasping platform with an simple off-the-shelf gripper, and conduct comprehensive simulation and real-world experiments to validate the effectiveness of our approach.
\end{enumerate}
\section{Related Work} \label{sec:relatedworks}

\subsection{{Model-based Aerial Grasping}}
Aerial grasping has been widely explored using classic model-based approaches. 
% Some studies focused on hardware design, especially the gripper mechanism \cite{fishman_dynamic_2021, appius_raptor_2022}. Others tackle perception challenges such as onboard self-localization and target pose estimation \cite{ bauer_open-source_2024, bauer_autonomous_2023, Ubellacker2024}. 
Most follow a sequential autonomy pipeline that consists of perception, planning, and control. 
The work \cite{fishman_dynamic_2021} presented the first prototype of a soft drone, equipped with a tendon-actuated soft gripper for aerial grasping. This physical system achieved a grasp success rate of 91.7\%, though its grasping speed was limited to 0.2~m/s. Subsequent work, RAPTOR \cite{appius_raptor_2022}, which used a custom Fin Ray\textsuperscript{\textregistered} \cite{crooks2016fin} soft gripper for more flexible grasping, represented a further step forward by achieving an 83\% grasp success rate with higher grasping speeds of 1~m/s.
% More recently, impressive results have been achieved even under fully onboard vision-based settings. For instance, 
Recently, Ubellacker et al.\cite{Ubellacker2024} demonstrated the fastest vision-based grasp reported to date, achieving grasping speeds of up to 2.0~m/s.
Their gripper adopted a complex structure consisting of four cable-actuated foam fingers, each driven by a DC motor. To control the gripper, they developed an optimal controller based on a finite element model of the fingers.

These classic pipelines typically decompose aerial grasping into two
separate modules: flight and gripper control \cite{bauer_open-source_2024, fishman_dynamic_2021, Ubellacker2024, appius_raptor_2022}. Such a modular approach often makes the timing of gripper closure dependent on manual experience. Their carefully designed soft grippers improve tolerance to flight control errors and provide contact cushioning during grasping \cite{fishman_dynamic_2021, Ubellacker2024}. However, they often come with relatively complex mechanical structures and may require dedicated controllers for actuation, increasing system complexity and reducing reproducibility. 
In contrast, our approach unifies precise flight control and active gripper control into a single lightweight policy, which is able to determine
the appropriate timing for both opening and closing the
gripper to enable seamless grasping during flight. Our aerial grasping platform uses a simple off-the-shelf gripper, directly actuated by servo angle control. Moreover, our two-stage DRL-based framework is flexible and has the potential to be extended to vision-based end-to-end aerial grasping.
%A unified policy could potentially replace the complex classical pipeline, improving both agility and robustness, as demonstrated in prior learning-based methods.
%In contrast, we address aerial grasping from a task-level perspective, and explore a unified learning-based approach that integrates flight and gripper control using a simple multi-layer perceptron (MLP) policy. This approach offers a more lightweight, efficient, and flexible solution.

\vspace{-0.3cm}
\subsection{{DRL-based Grasping}}
DRL has demonstrated remarkable performance in the field of autonomous flight. Notably, a DRL-based autonomous flight system \cite{kaufmann_champion-level_2023} has outperformed three human world champion pilots in real-world drone races. This remarkable work represents a significant milestone in mobile robotics and machine intelligence. However, these achievements primarily remain at the level of ``observing", without achieving active ``contacting" with the environment. In contrast, aerial grasping is a typical task for UAVs to interact with the environment, which inherently couples precise navigation with aerial manipulation, making it even more complicated and challenging.
%Although DRL-based aerial grasping has not yet been well-explored, 

DRL has made significant progress in ground-based robotic grasping. Huang et al. \cite{huang_earl_2023} proposed a novel DRL-based approach for dynamic grasping of unpredictably moving objects using a robotic arm. Hu et al. \cite{hu_grasping_2023} focused on a more challenging task, the grasping of living objects. %Their approach not only counteracts the living object's adversarial behavior but also ensures minimal contact force to prevent harm. 
Another work \cite{orsula_learning_2022} developed a DRL-based grasping policy using 3D octree observations, and achieved zero-shot sim-to-real transfer on a robotic arm mounted on a rover. 

While these ground-based approaches offer valuable insights for aerial grasping, transferring such methods to aerial platforms is nontrivial. Compared to robotic arms, quadrotors struggle with precise position control due to their underactuated dynamics, sensitivity to aerodynamic disturbances, and difficulty in accurate dynamics modeling. These factors together make DRL-based aerial grasping particularly challenging. Recent work \cite{dimmig_nonprehensile_2024} demonstrated aerial manipulation using model-based DRL, but it focused solely on non-prehensile pushing and did not achieve sim-to-real transfer. 
\section{Problem Statement} \label{sec:preliminaries}

\begin{figure}[h]
\centering
\includegraphics[width=0.47\textwidth,trim={0 2cm 0 3cm},clip]{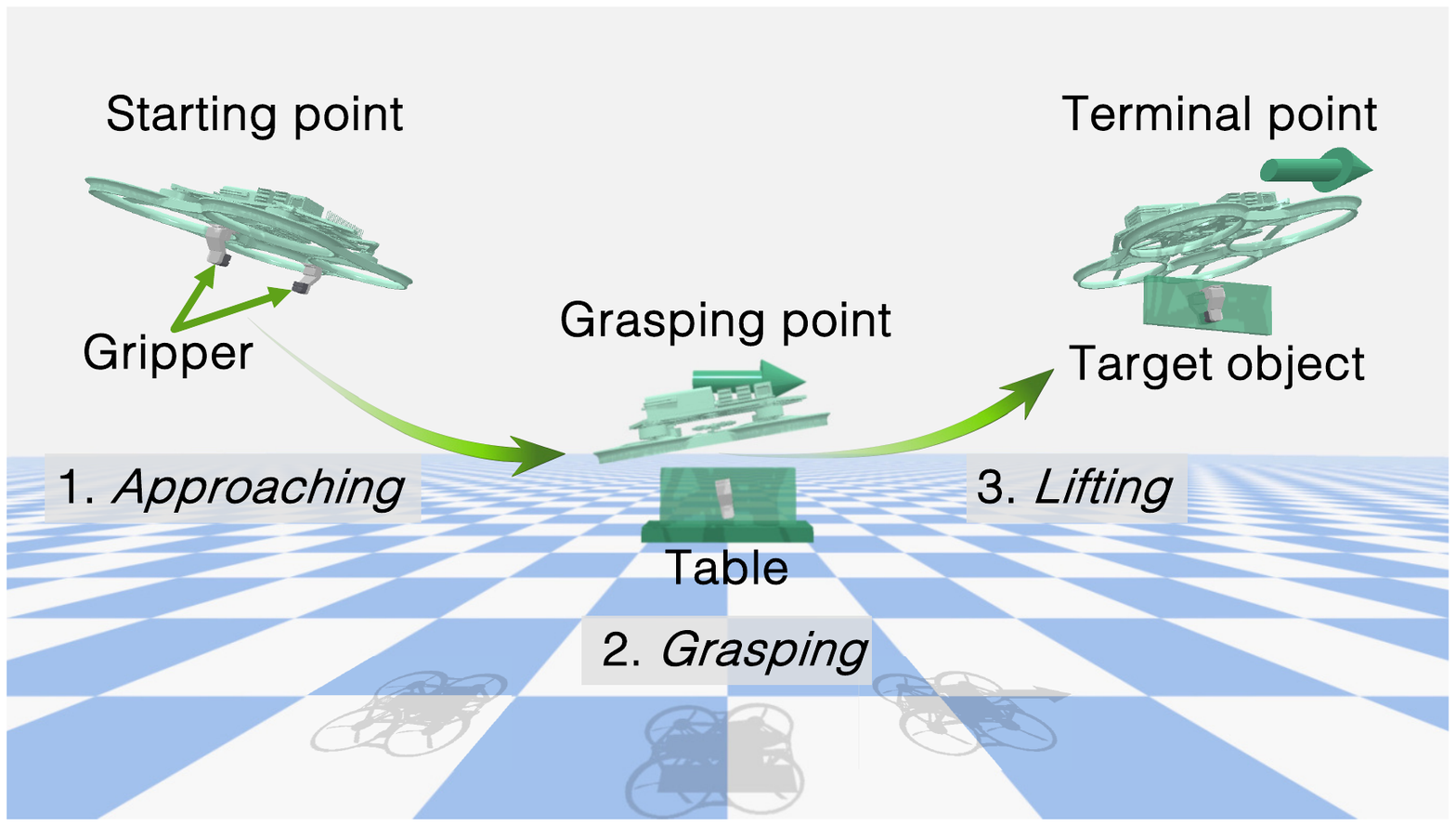} % trim order: left bottom right top
\caption{The simulation environment and the illustration of the aerial grasping process.}
\label{fig:grasp_env}
\end{figure}

\emph{High-speed aerial grasping} focuses on seamlessly grasping objects during high-speed flight, rather than slowing down significantly or hovering to execute the grasp. As shown in \Fig{fig:grasp_env}, we decompose the aerial grasping process into three sequential phases: (i) \textit{Approaching}, during which the quadrotor flies from the starting point to the grasping point; (ii) \textit{Grasping}, where the quadrotor closes its gripper to capture the object; (iii) \textit{Lifting}, in which the quadrotor transports the grasped object to the terminal point. Note that the decomposition is purely for explanatory purposes, while the actual grasping process is continuous and seamless.

\vspace{-0.2cm}

High-speed aerial grasping presents significant challenges. 
First, the flight control policy is required to be precise. %To validate the precision of our policy, 
To simplify the implementation, we employ a simple two-finger gripper (as shown in \Fig{fig:hardware}) with a maximum opening width of only 12~cm to grasp a target object of comparable size, measuring 2$\times$8$\times$14~cm$^3$. Consequently, the position error tolerance for flight control is constrained to approximately $\pm$ 5 cm along each axis at the moment of grasping.

Second,  the flight and gripper control must be responsive. Grasping during high-speed flight requires not only accurate timing for closing the gripper, but also a fast-responding gripper actuation mechanism.  For example, if the quadrotor is flying at 1.5~m/s, even a 0.1s shift in gripper actuation would cause a 15~cm drift, enough to miss the object (2$\times$8$\times$14~cm$^3$ in our setting) or cause a collision. 

\begin{figure}[h]
\centering
\includegraphics[width=0.33\textwidth,trim={3cm 6.2cm 3cm 6.5cm},clip]{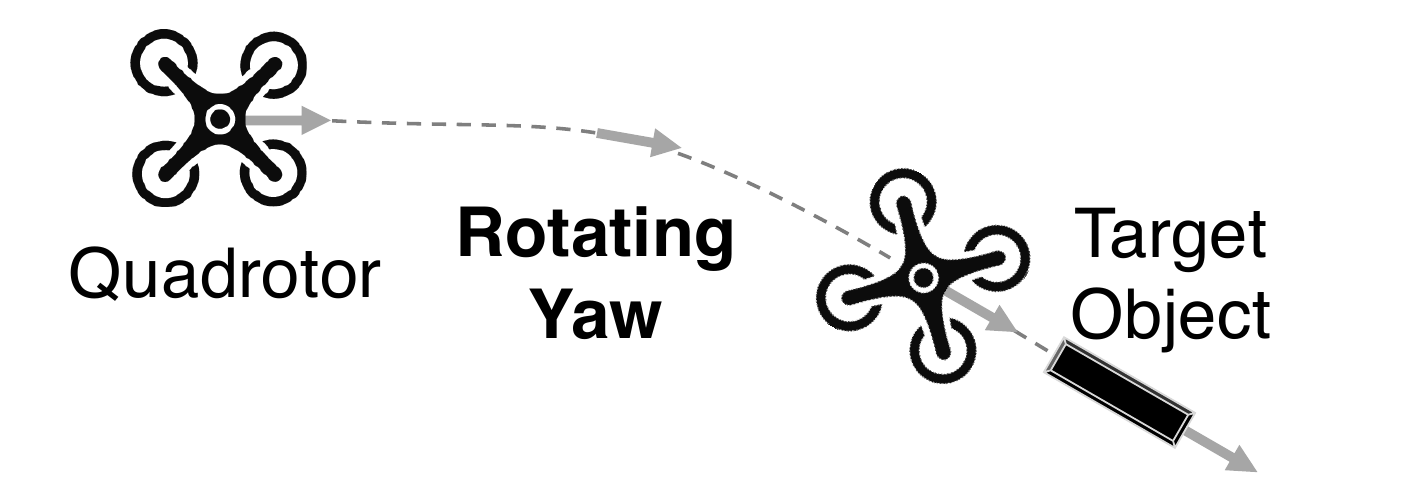} % trim order: left bottom right top
\caption{To grasp a target object placed at an arbitrary position and orientation, the quadrotor has to rotate its yaw angle to align with that of the object during the \textit{approaching} phase.}
\label{fig:rotating_yaw}
\end{figure}

Third, objects placed at different positions and orientations require different grasp poses. The desired grasp pose is assumed to be known in advance, represented by the target object's yaw angle. Given our simple two-finger gripper mechanism (as shown in \Fig{fig:hardware}), the quadrotor is required to align its yaw angle with that of the target object during the \textit{approaching} phase to enable a rapid grasp while swooping over the object, as illustrated in \Fig{fig:rotating_yaw}. 
% Since the gripper is symmetric, the yaw angle of the target object only ranges from -90 to 90 degree. 
While few prior works focus on yaw control, we argue that it enables more versatile aerial grasping. By explicitly incorporating yaw control into the learning process, the trained policy is able to grasp objects in a wider range of poses without relying on sophisticated soft grippers.
\vspace{-0.3cm}

Finally, there exists a conflict between learning flight control and gripper control simultaneously, which makes training an aerial grasping policy directly from scratch particularly  challenging. In the early stage of training, the agent learns to fly to the grasping point and perform grasping.  However, since the position control is still inaccurate and the agent has not yet mastered proper gripper control, premature grasp attempts often cause  the gripper to knock over the object. This triggers penalties and leads to early episode termination, which in turn hinders the exploration necessary for learning flight control.

\section{Methodology} \label{sec:methods}

\begin{figure*}[ht]
\centering
\includegraphics[width=1.0\textwidth, trim={0.2cm 5.0cm 0.2cm 5.0cm},clip]{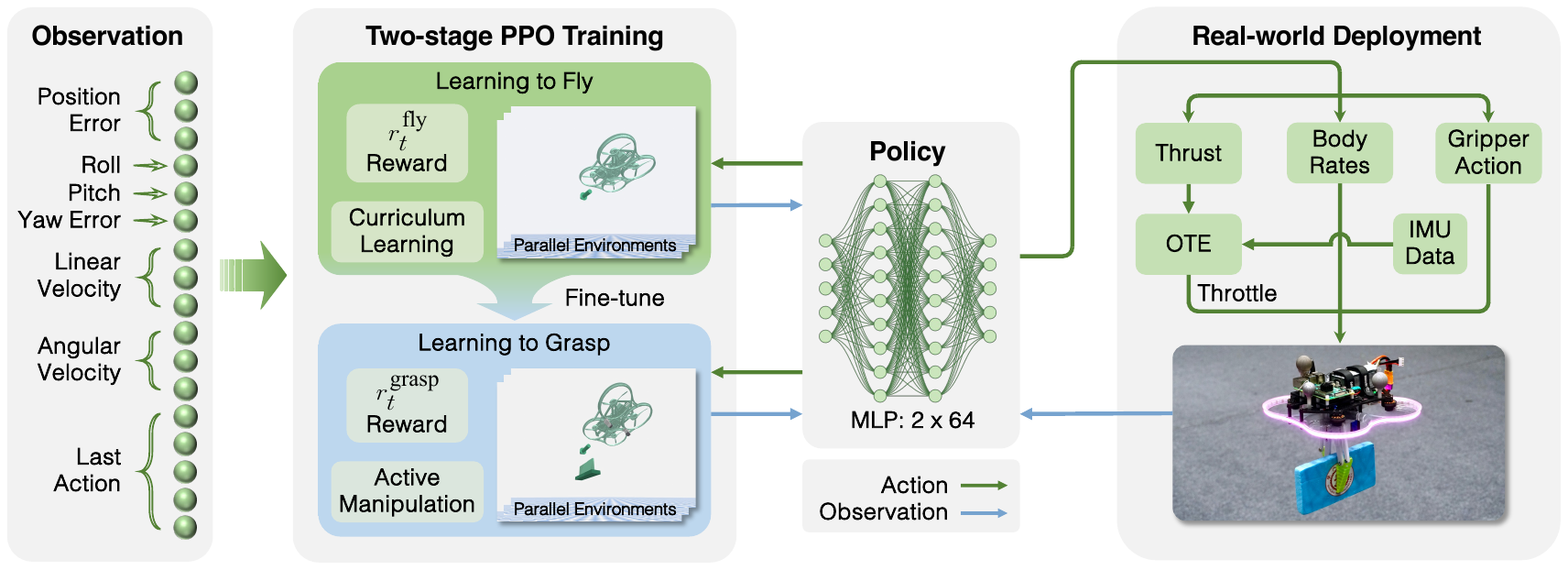} % trim order: left bottom right top
\caption{System overview. Our two-stage DRL-based approach first trains a flight control policy and then fine-tunes it to acquire gripper control, finally yielding a unified and lightweight aerial grasping policy. Each stage is guided by a tailored reward function. The policy network takes the current and desired states of the quadrotor as input, and outputs a CTBR command for flight control and a gripper control command. OTE refers to the Online Throttle Estimation module in Section~\ref{subsec:real_experiments}.} 
\vspace{-0.5cm}
\label{fig:algo}
\end{figure*}

Our DRL-based approach for high-speed aerial grasping consists of two stages: \emph{Learning-to-fly} and \emph{Learning-to-grasp}. The \emph{Learning-to-fly} stage trains a policy for set-point flight, which is then fine-tuned to acquire grasping skills in the \emph{Learning-to-grasp} stage. Finally, the trained policy seamlessly integrates both flight control and gripper control to achieve high-speed aerial grasping. The overview of our approach is shown in \Fig{fig:algo}.

\vspace{-0.3cm}
\subsection{Policy Definition}

The policies to be trained in the two stages share the same structure: a multi-layer perceptron (MLP) that maps an observation $\mathbf{o}_t$ to an action $\mathbf{a}_t$.

\textbf{Observation:} The observation includes the quadrotor's state and the desired pose, which is defined as:
\begin{equation}
\small
    \mb{o}_t = \{\mb{p}_t - \mb{p}_\text{d}, \phi_t, \theta_t, \psi_t - \psi_\text{d}, \mb{v}_t, \bs{\omega}_t, \mb{a}_{t-1}\}\in\mathbb{R}^{17}
    \label{eq:obs}
\end{equation}
 where $\mathbf{p}_t$, $\mathbf{v}_t$, and $\boldsymbol{\omega}_t$ denote the quadrotor's position, linear velocity, and angular velocity vectors, respectively. $\phi_t$, $\theta_t$, and $\psi_t$ represent the roll, pitch, and yaw angles of the quadrotor. $\mb{p}_\text{d}$ and $\psi_\text{d}$ are the desired position vector and yaw angle, while $\mb{a}_{t-1}$ is the action applied in the last step. Note that $\boldsymbol{\omega}_t$ is represented in the body frame, while the other states are given in the world frame. %We use Euler angles for intuitive orientation representation. It is worth noting that the aerial grasping task does not involve high-agility maneuvers, thus eliminating the concern regarding the ambiguity of Euler angle representation.

\textbf{Action:} The normalized action $\mb{a}_t \in [-1,1]^5$ consists of a flight control command $\mb{a}_t^{\text{fly}} \in [-1,1]^4$ in the form of the Collective Thrust and Body Rates (CTBR) \cite{Kaufmann_benchmark_2022}, and a gripper control command $a_t^{\text{gp}} \in [-1,1]$.
Due to its ability to enable agile maneuvers, the CTBR control modality is widely adopted in drone racing \cite{song_reaching_2023, penicka_learning_2022, kaufmann_champion-level_2023}. Moreover, it has shown high robustness to dynamics mismatch and ensures reliable sim-to-real transfer \cite{Kaufmann_benchmark_2022}. $a_t^{\text{gp}}$ is designed as a continuous variable, which enables more fine-grained gripper control than a discrete one. Actually, it denotes the normalized gripper width, where $-1.0$ represents a fully closed state and $1.0$ indicates a fully open one. Note that $a_t^{\text{gp}}$ is not used in the \emph{Learning-to-fly} stage. It is included solely to preserve the output structure of the policy network for the subsequent training of \textit{Learning-to-grasp}.

\vspace{-1mm} % important for margin
\subsection{Learning to Fly}
\textbf{Environment setting:} The training environment is an object-free 3D space where the quadrotor has to fly from a starting point to a target position while aligning its yaw with a desired angle. This setting is designed to meet the requirement of grasping objects placed at different positions and orientations.
%This setup reflects the challenges of aerial dynamic grasping, where the target object is randomly placed and the quadrotor is typically moving relative to it at the moment of grasping. To enable rapid grasping, the quadrotor must align its heading (yaw) orientation with that of the target object, enabling it to sweep over and capture the object, much like an eagle swooping down to catch its prey.

\textbf{Reward function:} The reward function for the flight task at time step $t$ is shaped as follows:
\begin{equation}
\small
\label{eq:fly_reward}
r_t^{\text{fly}} = r_t^{\mb{p}} + r_t^{\psi} + r_t^{\text{sm}} + r_t^{\text{safe}}
\end{equation}
where $r_t^{\mb{p}}$ encourages the quadrotor to approach the desired position, while $r_t^{\psi}$ guides the quadrotor to align its yaw angle with the desired one. $r_t^{\text{sm}}$ rewards smooth actions for stable flight. $r_t^{\text{safe}}$ serves as a penalty that is activated only when the quadrotor violates safety constraints, such as exiting the predefined bounding box or tilting excessively. Once $r_t^{\text{safe}}$ is triggered, the episode is terminated early. Specifically, these reward terms are formulated as follows, where $\mathbb{I}\{ \cdot \}$ is the indicator function: 
\begin{equation}
\small
\begin{alignedat}{2}
    &r_t^{\mb{p}}      &&= - \lambda_1 \|\mb{p}_t - \mb{p}_\text{d}\| \\
    &r_t^{\psi}        &&= 1.0 - \exp{(\lambda_2 \|\psi_t - \psi_\text{d}\|^2)} \\
    &r_t^{\text{sm}}   &&= - \lambda_3 \|\bs{\omega}_t\|^2 - \lambda_4 \|\mb{a}_t^{\text{fly}} - \mb{a}_{t-1}^{\text{fly}}\|\\ 
    &r_t^{\text{safe}} &&= -10.0 \cdot \mathbb{I} \{\text{safety constraints violated}\} 
\end{alignedat}
\end{equation}
\vspace{-1mm} % important for margin

% \textbf{Termination condition:} An episode terminates when any of the following occurs: (i)
% $\|\mb{p}_t - \mb{p}_\text{d}\|$ and $\|\psi_t - \psi_\text{d}\|$ fall within predefined thresholds, respectively, which is also considered a successful set-point tracking, (ii) the quadrotor exits a predefined bounding box or its attitude becomes excessively tilted, (iii)the episode time limit is hit. 

\subsection{Learning to Grasp}

\textbf{Environment setting:} The training environment is shown in \Fig{fig:grasp_env}.
A trial is considered successful if the quadrotor grasps the object and transports it to the terminal point. Note that during the \textit{approaching} phase, $\mb{p}_\text{d}$ is initialized to a suitable grasping position above the target object, and $\psi_\text{d}$ is set to the object's yaw angle. Once the object is grasped, the process immediately transitions to the \textit{lifting} phase, where $\mb{p}_\text{d}$ and $\psi_\text{d}$ are  switched to those of the selected terminal point. 

\textbf{Reward function:}
For the training of grasping, we carefully design a composite reward function $r_t^{\text{grasp}}$ that combines sparse and dense rewards as follows:
\begin{equation}
\small
\begin{alignedat}{2}
    &r_t^{\text{grasp}}    &&= r_t^{\text{phase}} + r_t^{\text{gp\_instr}} + r_t^{\text{gp\_sm}} + r_t^{\text{crash}} \\
    &r_t^{\text{phase}}    &&= \lambda_5^{i} \cdot \mathbb{I} \{\text{phase $i$ first completed}\} \\
    &r_t^{\text{gp\_instr}}&&= \lambda_6 \cdot a_t^{\text{gp}} \cdot (0.5 - \mathbb{I} \{\text{\textit{approaching} completed}\}) \\
    &r_t^{\text{gp\_sm}}    &&= - \lambda_7 \cdot \|a_t^{\text{gp}} - a_{t-1}^{\text{gp}}\| \\
    &r_t^{\text{crash}}     &&= -10.0 \cdot \mathbb{I} \{\text{safety constraints violated}\}
\end{alignedat}
\label{eq:grasp_reward}
\end{equation}

\begin{itemize}
    \item Phase reward $r_t^{\text{phase}}$: Following \cite{orsula_learning_2022}, we adopted a hierarchical flow reward $r_t^{\text{phase}}$ that integrates sparse rewards from the three sequential phases. Each phase provides a reward only at the time step upon its completion. The reward increases exponentially with phase order $i$ as $\lambda_5^{i}$, where $i \in \{1,2,3\}$. The completion conditions are: (i) \textit{Approaching}: $\|\mb{p}_t - \mb{p}_\text{d}\| < \delta_{\text{dist}} \ \text{and} \ \|\psi_t - \psi_\text{d}\| < \delta_\psi$, where $\delta_{\text{dist}}>0$ and $\delta_\psi>0$ denote the distance and yaw angle thresholds;
    (ii) \textit{Grasping}: the two gripper fingers collide with the object from opposite directions, indicating a successful grasp;
    (iii) \textit{Lifting}: $\|\mb{p}_t - \mb{p}_\text{d}\| < \delta_{\text{dist}}\ \text{and}\ \|\psi_t - \psi_\text{d}\| < \delta_\psi\ \text{and}\ \|\mb{p}_t^{\text{obj}} - \mb{p}_\text{d}\| < \delta_{\text{obj\_dist}} $, where $\mb{p}_t^{\text{obj}}$ is the object position and $\delta_{\text{obj\_dist}}>0$ is also a distance threshold. This reward term guides the agent to master the process of aerial grasping.
    \item Gripper instruction reward $r_t^{\text{gp\_instr}}$: Inspired by \cite{shahid_learning_2020}, we design this reward term to instruct the gripper's behaviors. This term encourages the agent to open the gripper in preparation for grasping during the \textit{approaching} phase by guiding $a_t^{\text{gp}}$ toward 1.0. It further guides the agent to close the gripper upon reaching the grasping point and to keep it closed until the object is transported to the terminal point by driving $a_t^{\text{gp}}$ toward -1.0.
    \item Gripper smoothness reward $r_t^{\text{gp\_sm}}$: This term rewards smooth gripper actions, which helps prevent the grasped object from slipping out of the gripper due to the gripper jitter and reduces energy consumption resulting from unnecessary gripper operations.
    \item Crash reward $r_t^{\text{crash}}$: This is a penalty activated only when the quadrotor violates the safety constraints by exiting the predefined bounding box, tilting excessively, or colliding with the table or the object. Once $r_t^{\text{crash}}$ is triggered, the episode is truncated early.
\end{itemize}

It is worth noting that during training, we do not explicitly specify the grasping speed (i.e., the speed of the quadrotor at the moment of grasping). Instead, the agent learns to control its flight speed autonomously as a result of reward optimization. Specifically, the policy balances between approaching the target quickly and ensuring a successful grasp through the designed reward terms.

\subsection{Policy Training} \label{sec:policy_training}
 % PPO is a model-free policy gradient method designed to optimize policies efficiently while avoiding performance collapse by ensuring new policies remain close to the old ones. PPO has gained widespread popularity in the filed of robotics due to its efficiency, simplicity, and effectiveness in handling complex continuous control problems.
We employ the Proximal Policy Optimization (PPO) algorithm \cite{schulman2017proximal} to train our policy for both \textit{Learning-to-fly} and \textit{Learning-to-grasp}. Our policy network and value network share the same structure, consisting of an MLP with 2 hidden layers, each containing 64 neurons. The activation function used throughout the network is Tanh, except for the output layer, which employs a linear activation function. The final output of the policy network is then clipped within the range of $[-1,1]$. We set the hyperparameters of the reward terms as follows: $\lambda_1 = 1.0$, $\lambda_2 = 0.1$, $\lambda_3 = 0.1$, $\lambda_4 = 0.4$, $\lambda_5 = 10.0$, $\lambda_6 = 4.0$, and $\lambda_7 = 5.0$.

For the training of \textit{Learning-to-fly}, we design a curriculum learning procedure for yaw control to accelerate convergence. Specifically, we start by aligning the quadrotor's initial yaw angle with $\psi_\text{d}$, allowing the agent to focus solely on learning position control without concern of controlling its yaw. Once the agent has achieved reliable position following, we gradually increase the error between its initial yaw angle and $\psi_\text{d}$ at the beginning of each episode, driving the agent to acquire yaw control progressively.
\section{Experiments} \label{sec:experiments}

In this section, we conduct a series of simulation and real-world experiments to answer the following questions:
\begin{enumerate}
\item How does our two-stage learning approach compare to training from scratch in terms of sample efficiency and final performance?

\item Are all the components of our reward functions valid and necessary?

\item What are the effects of grasping speed and relative object yaw angle on grasp success rate?

\item How well does the trained policy perform when deployed on a real platform  without any fine-tuning?

\item Can our real platform robustly grasp different objects?
\end{enumerate}

\subsection{Simulation Experiment Setup}

Our customized simulation environment is built upon \textit{gym-pybullet-drones} \cite{panerati_learning_2021}, an open-source OpenAI gym-style quadcopter simulator based on the Bullet physics engine~\cite{coumanspybullet}. Both training and evaluation are conducted on a desktop with an Intel Core i9-11900K CPU and an Nvidia RTX 3060 GPU. For efficient implementation, we use Stable-Baselines3~\cite{stable-baselines3} to deploy the PPO algorithm. Moreover, we simulate multiple parallel environments to collect data for speeding up training. With the carefully designed reward functions and training framework, the entire two-stage training process takes less than 60 minutes.

\subsection{Comparison with Training From Scratch}
\label{subsec:comparison_from_scratch}

Our two-stage training pipeline fine-tunes the pretrained flight control policy to acquire aerial grasping maneuvers. A natural question arises: is it possible to achieve comparable performance through training from scratch (TFS)? To investigate this, we compared the learning curves of our approach with those of TFS. 
For the implementation of TFS, we used the \textit{Learning-to-grasp} environment and  the reward function of TFS is defined as a combination of $r_t^{\text{fly}}$ and $r_t^{\text{grasp}}$. The curriculum learning procedure for yaw control in Section~\ref{sec:policy_training} is also applied during training. We also made careful efforts to retune the weights of individual reward terms to ensure a fair comparison. Note that all other settings are consistent with our pipeline.
% For the implementation of TFS, we used the \textit{Learning-to-grasp} environment and incorporated the settings from the \textit{Learning-to-fly} stage, including the reward function $r_t^{\text{fly}}$ (\ref{eq:fly_reward}) and the curriculum learning procedure in Section~\ref{sec:policy_training}. Note that the other settings are consistent with our pipeline. 

The success rate (SR) curves in \Fig{fig:training_result} clearly illustrate that Swooper achieves robust flight and grasping policies with high sample efficiency, while TFS fails, even consistently maintaining a 0\% SR. During the evaluation of TFS, we observed that the agent converged to a suboptimal hovering policy and even failed to achieve reliable set-point following. We ascribe this to the conflict between learning flight and gripper control simultaneously. The agent in TFS learned gripper control before mastering stable flight control, which resulted in frequent object knock-downs and crashes, triggering the crash reward and further hindering attempts at set-point following. 
% We ascribe this to learning gripper control without mastering stable flight control. This results in frequent object knockdowns and crashes, triggering the crash reward and further hindering attempts at set-point following. 
Moreover, the reward function becomes highly complex when the flight and grasping tasks are coupled, making hyperparameter tuning even more difficult.
% Thus, the success of our method can be attributed to the incremental learning strategy, where pretraining on stable flight allows the agent to first acquire the foundational skills needed for precise control, which in turn enables effective grasping. By contrast, training from scratch, with its complex reward function and the difficulty of parameter tuning, fails to decompose the task into manageable sub-goals, which inhibits the learning process and leads to failure in both grasping and flight stabilization.

\subsection{Ablation Studies} 
We conducted ablation studies to validate the effectiveness of the tailored reward components, specifically focusing on the phase reward $r^{\text{phase}}$ and the gripper instruction reward $r^{\text{gp\_instr}}$. To ensure validity, these ablation studies were conducted by fine-tuning the same trained flight control policies.

\begin{figure}[h]
\centering
\includegraphics[width=0.48\textwidth,trim={3.8cm 2.5cm 3.8cm 2.5cm},clip]{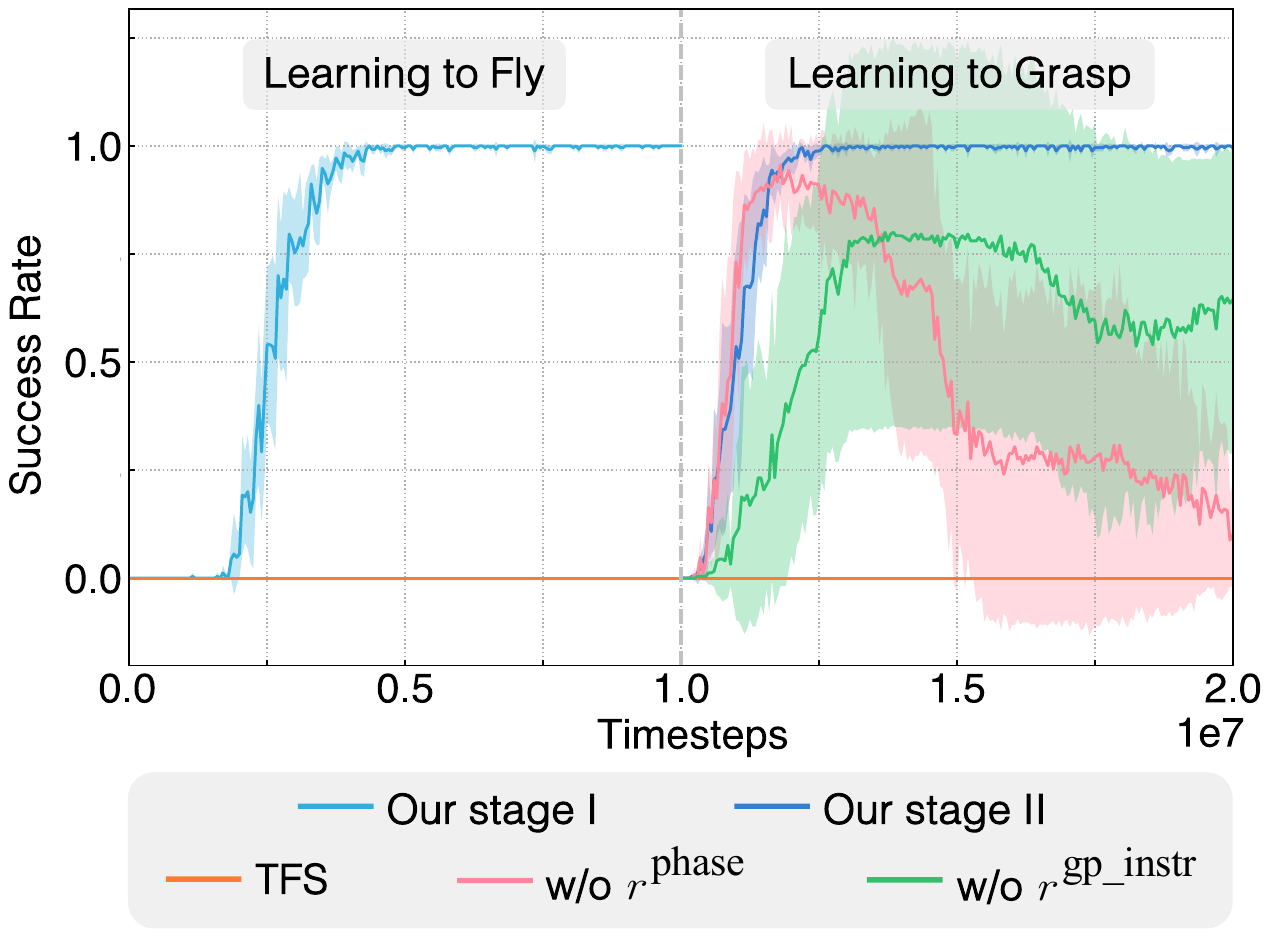} % trim order: left bottom right top
\caption{Learning curves of policies trained with different settings. Each solid line indicates the mean performance across 5 training runs with different random seeds, and its shaded band represents the standard deviation. Note that the criteria for success differ between the two stages and the success rate of TFS refers to the grasp success rate. 
\vspace{-0.3cm}
}
\label{fig:training_result}
\end{figure}

\textbf{$r^{\text{phase}}$}: As shown in \Fig{fig:training_result}, removing $r^{\text{phase}}$ results in unstable training and significant performance degradation. As training progressed, the quadrotor gradually failed to reach the terminal point in the \textit{lifting} phase and often drifted aimlessly. We attribute this to the absence of the phase reward, which guides the agent to master the grasping process step by step. Without this guidance, the policy suffers from catastrophic forgetting, gradually losing its ability to complete the entire aerial grasping process.

\textbf{$r^{\text{gp\_instr}}$}: The learning curve in \Fig{fig:training_result} shows significant instability and relatively low SR when $r^{\text{gp\_instr}}$ is absent. Thanks to the guidance of the phase reward, the agent was still able to acquire basic gripper operation skills even without the instruction from $r^{\text{gp\_instr}}$. The primary cause of failures was that the gripper often failed to open until reaching the grasping point, leading to gripper-object collisions. This result emphasizes the importance of the gripper instruction reward in guiding the agent to master the appropriate timing for both opening and closing the gripper. Interestingly, when $r^{\text{gp\_instr}}$ is applied, although it rewards closing the gripper only after reaching the grasping point, the agent implicitly learns to close the gripper slightly before reaching the grasping point, thus enabling seamless and rapid aerial grasping.

\subsection{Correlation Analysis of Grasping Performance}
\begin{figure}[h]
\centering
\includegraphics[width=0.48\textwidth,trim={0cm 0cm 0cm 0cm},clip]{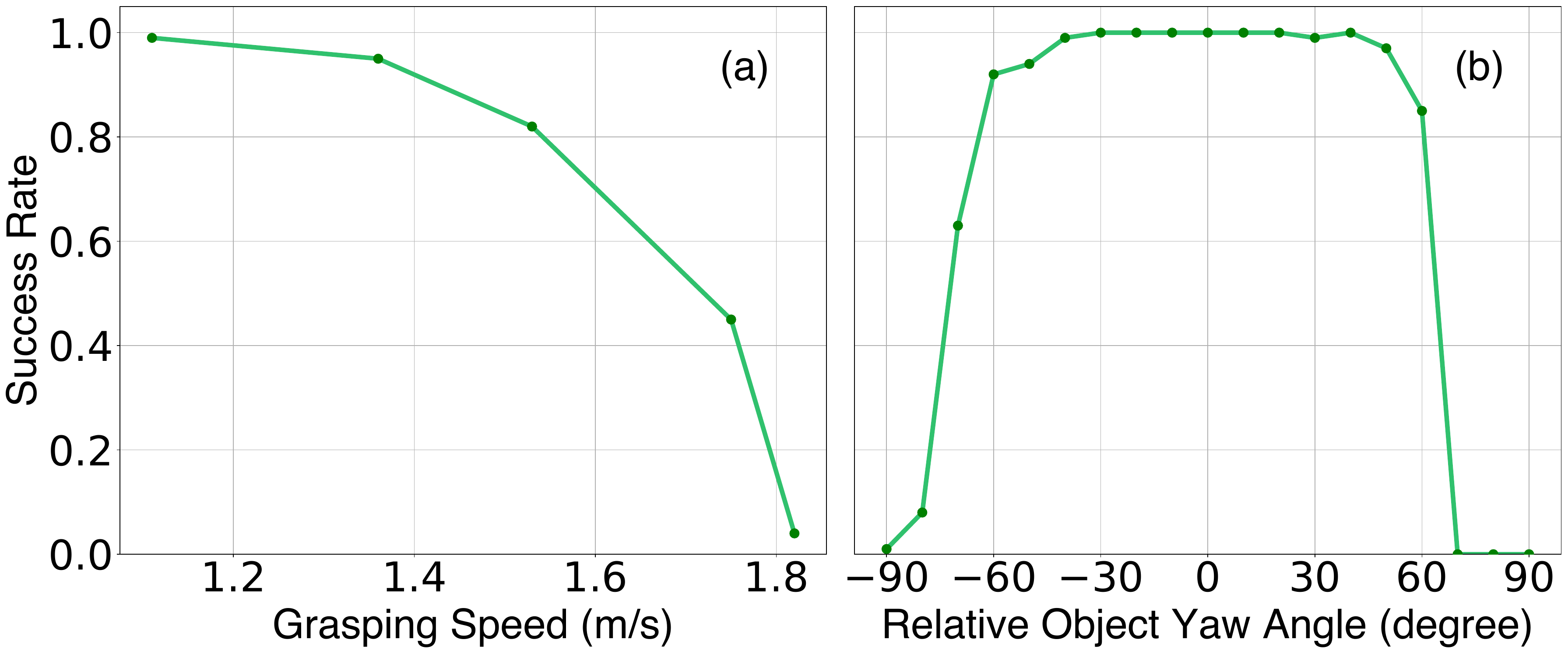} % trim order: left bottom right top figures/frame.pdf / frame_mask.pdf
\caption{Grasp success rate vs. grasping speed and relative object yaw angle. The relative object yaw angle also refers to the initial yaw error between the quadrotor and the object.}
\label{fig:SrVsSpeedYaw}
\end{figure}

We conducted a set of experiments to investigate the effects of grasping speed and relative object yaw angle on grasping performance, and to evaluate the limits of our policy. \Fig{fig:SrVsSpeedYaw}~(a) indicates that success rate decreases as grasping speed increases. Our policy remains robust up to approximately 1.5~m/s, achieving over 80\% success rates. Beyond this point, performance degrades rapidly, approaching zero at 1.82 m/s. This trend can be clearly explained, as higher flight speeds inherently increase the difficulty of maintaining precise pose control and triggering the gripper at the correct timing.

As shown in \Fig{fig:SrVsSpeedYaw}~(b), the policy demonstrates high grasp success rates for relative object yaw angles between $-60^{\circ}$ and $60^{\circ}$, exhibiting a nearly symmetric performance curve. Beyond this range, the success rate degrades significantly. 
This degradation mainly arises from two factors: (i) during training, the relative object yaw angles are initialized within the range of $-60^{\circ}$ to $60^{\circ}$, which constrains generalization beyond this interval, and (ii) larger yaw angles make the task more challenging. To perform grasping, the quadrotor has to rotate its yaw while approaching the object rapidly. We observed that when the quadrotor performs a large yaw rotation, lateral drift tends to intensify during flight, often causing it to approach the object from the side and the gripper to topple the object.
% Given our two-finger gripper mechanism, successful grasping requires the quadrotor to align its forward direction with the object’s yaw before reaching the grasping point, as illustrated in \Fig{fig:rotating_yaw}. This requires the quadrotor to learn to plan smooth, curved trajectories; However, our relatively simple reward design is insufficient to guide the agent to acquire such complex trajectory planning behaviors. Since the quadrotor tends to follow straight-line trajectories, it often approaches the object from the side, causing the gripper to strike and topple the object.

\vspace{-0.5cm}
\subsection{Real-world Experiments} \label{subsec:real_experiments}
\begin{figure}[h]
\centering
\includegraphics[width=0.46\textwidth,trim={0.5cm 2.0cm 0.5cm 2.5cm},clip]{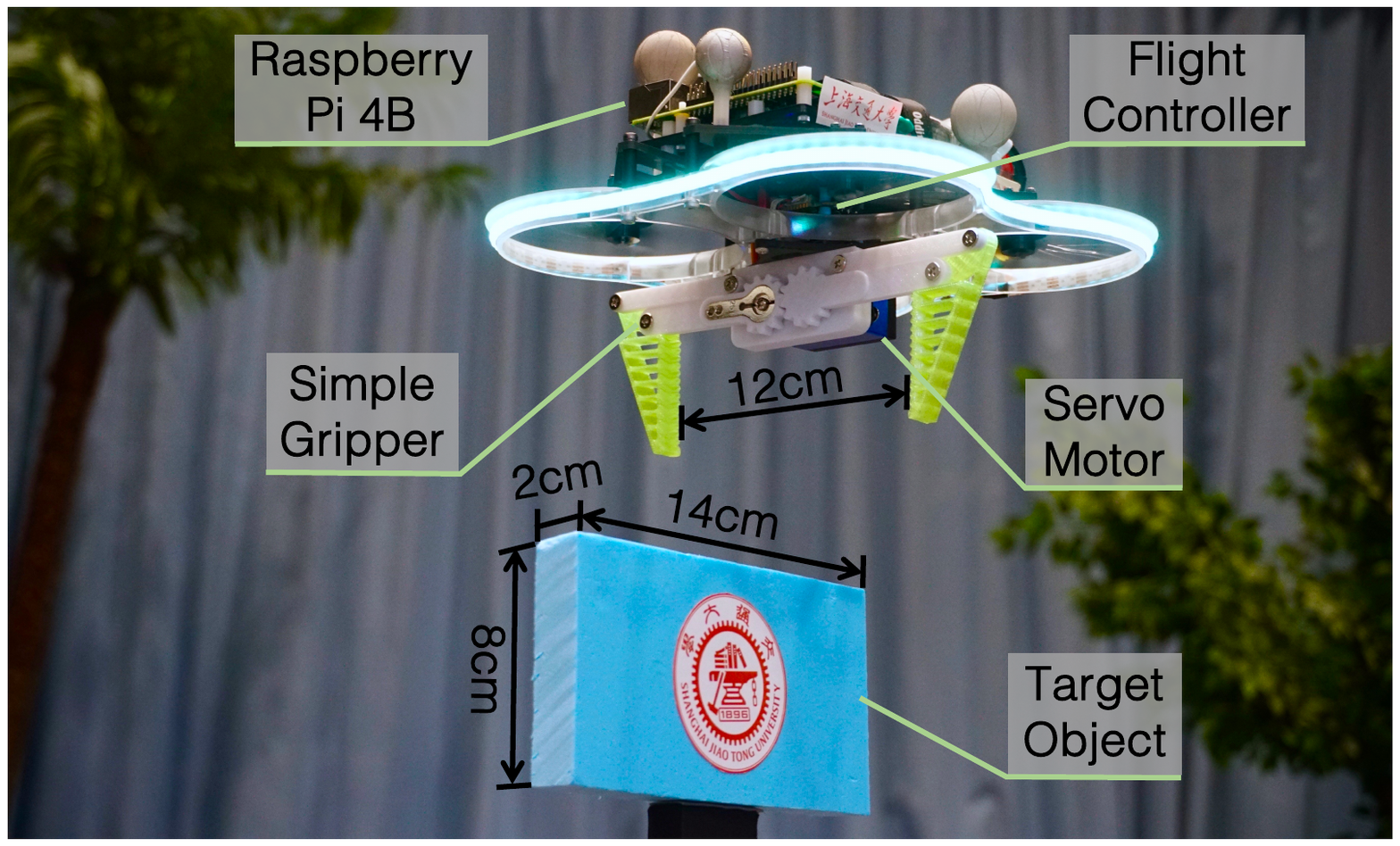} 
\caption{The customized aerial grasping platform and the target object (a rigid foam block weighing 7g) used in real-world experiments.}

\label{fig:hardware}
\end{figure}

\textbf{1) Hardware setup:} Our hardware setup consists of two main components: a quadrotor and a simple gripper, as shown in \Fig{fig:hardware}.

The quadrotor platform is extensively customized to meet the requirements of the high-speed aerial grasping task. The quadrotor is constructed in an inverted configuration, using a flight frame designed by \cite{liu2024omninxt}, an Oddity RC XI35 protector, T-MOTOR F2004 3000KV motors, and GEMFAN D90S-3 3.5-inch propellers. For flight control, we use a compact FPV flight stack comprising a Mamba MK4 F722 MINI flight controller and a 4-in-1 electronic speed controller with a peak current capacity of 45A. A modified version of the BetaFlight firmware \cite{zhang_back_2024} is employed to facilitate autonomous flight. 
%The quadrotor is powered by a Tattu 2200mAh 4S 16.8V lithium battery with a 45C discharge rate, ensuring flight endurance and sufficient power for grasping. 
Furthermore, Raspberry Pi 4B, a low-cost and compact single-board computer, is selected as the onboard compute resource for the policy deployment. It features a quad-core 1.5GHz ARM Cortex-A72 processor, 8GB RAM, and extensive GPIO pins to facilitate the convenient implementation of the servo motor control. Thanks to our extremely lightweight policy, each inference on our resource-constrained tiny computer takes only about 1.0 ms.

%Due to the extreme lightweight nature of our network, inference reaches 10,000 Hz on this computation-constrained tiny computer.

For grasping, we employ an off-the-shelf 3D-printed gripper and mount it to the bottom of the quadrotor. It adopts a telescopic operating mechanism with a maximum opening width of 12~cm. The gripper is driven by an SPT 5828W servo motor, which is lightweight (67g) and highly responsive, enabling the gripper to close within 0.1s. 
In contrast to the carefully designed grippers with complex mechanical structures used in prior works \cite{fishman_dynamic_2021, Ubellacker2024, appius_raptor_2022, thomas_avian-inspired_2013}, our two-finger gripper is an off-the-shelf component, which is simple, lightweight, and low-cost.
Its opening and closing operations are directly actuated by the servo angle control, eliminating the need for additional complex controllers.
The aerial grasping platform weighs 598g totally and can produce a max thrust-to-weight ratio of 2.8, ensuring agility for high-speed grasping.
% For grasping operation, a 3D-printed gripper is mounted to the bottom of the quadrotor. It is driven by an SPT 5828W servo motor, which is lightweight (67g) and highly responsive, with a maximum speed of $\text{60}^\circ\text{/}\text{0.06}\,\text{s}$. These features make it well-suited for the high-speed aerial grasping task. The gripper features a telescopic operating mechanism, with a maximum opening width of 12~cm. The two soft fingers provides cushioning to reduce the interference caused by impulsive forces at grasping. In contrast to the carefully designed grippers with complex structures in prior works \cite{fishman_dynamic_2021, appius_raptor_2022, Ubellacker2024, thomas_avian-inspired_2013}, our gripper is simple and easy to implement. The aerial grasping platform weighs 598g totally and can produce a max thrust-to-weight ratio of 2.8.

\textbf{2) Implementation:} In the real-world experiments, the state information of both the quadrotor and the target object was obtained through the Vicon motion capture system, except for the angular velocity of the quadrotor, which was measured by the onboard IMU due to its smoother data.
The state information from both the Vicon system and the IMU was updated at about 150~Hz. 
The trained policy ran at 30~Hz to ensure stable operation under constrained computational resources. To bridge the sim-to-real gap, we introduce an Online Throttle Estimation (OTE) module \cite{wang_agile_2024}, which estimates the quadrotor’s thrust model from IMU data and converts thrust commands into throttle signals. This module helps compensate for dynamics mismatch and battery degradation during flight, and improve flight control performance.
% The trained policy ran on the onboard computer at a rate of 30~Hz, sending commands to the lower-level controller to achieve the desired control behavior. 

To validate the capability of the trained policy for performing rapid grasps across various positions and
orientations, we initialized the quadrotor at a fixed takeoff position and randomly placed the object at five different positions, each with a selected yaw angle, and conducted five consecutive trials for each configuration using the same policy. As illustrated in \Fig{fig:realflight}, the real-world aerial grasping process follows this sequence: the quadrotor first takes off and hovers at a preset height, then the policy is activated to perform the grasping, and finally, the quadrotor lands after the object is transported to the terminal point. The results of these tests are presented in Table \ref{tab:successRates}, where the object's position and yaw angle were measured relative to the quadrotor's pose at the moment of policy activation. 

\begin{table}[h]
\scriptsize
    \renewcommand{\arraystretch}{1.3}
    \caption{Grasping results across positions and yaw angles}
    \centering
    \begin{tabular}{@{\extracolsep{\fill}} c c c c c @{}}
        \toprule
        \makecell{Relative Object \\ Position [m]} & \makecell{Relative Object \\ Yaw Angle} & \makecell{Success \\ Rate} & \makecell{Avg. Grasping \\ Speed [m/s]} \\
        \midrule
        (0.85, 0.05, -0.15) & $0^{\circ}$ & 5/5 & 0.91  \\
        (0.95, -0.10, -0.15) & $-20^{\circ}$ & 5/5 & 1.16 \\
        (0.75, 0.15, -0.20) & $20^{\circ}$ & 4/5 & 0.92 \\
        (1.20, -0.50, -0.30) & $-40^{\circ}$ & 4/5 & 1.53 \\
        (1.05, 0.70, -0.40) & $40^{\circ}$ & 3/5 & 1.41 \\
        \midrule
        Total & & 21/25(84\%) &  1.19\\
        \bottomrule
    \end{tabular}
    \label{tab:successRates}
\end{table}

\begin{figure}[h]
\centering
\includegraphics[width=0.48\textwidth,trim={0.8cm 4.5cm 0.8cm 4.2cm},clip]{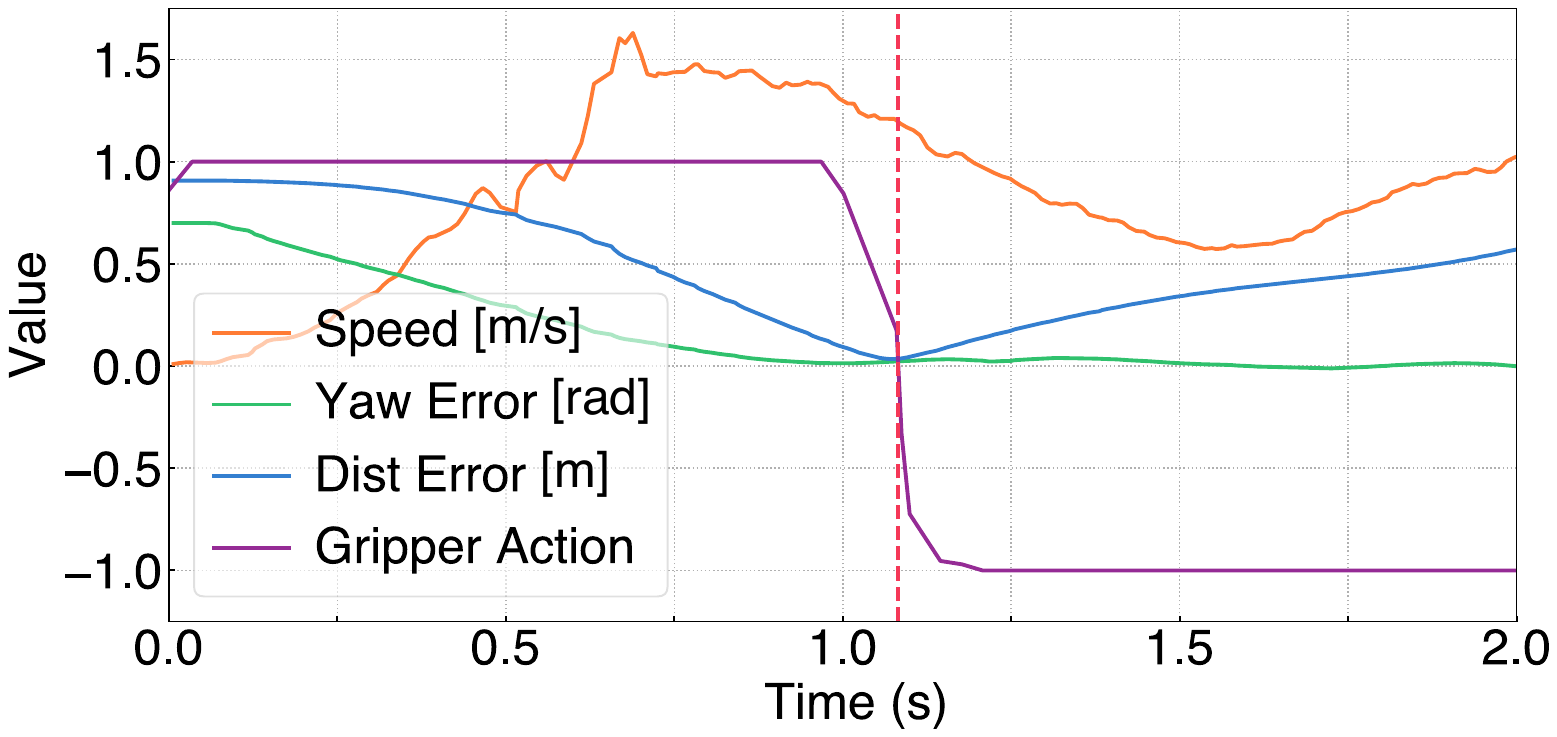} % trim order: left bottom right top
\caption{Time series of the quadrotor flight speed, yaw angle and distance errors relative to the desired  grasping point, and the gripper action during a real-world trial. The vertical dashed line indicates the moment of grasping.}
\label{fig:real_curve}
\end{figure}

Since we use a simple gripper with a maximum opening
width of 12 cm to grasp a target object of comparable size (measuring 2$\times$8$\times$14~cm$^3$, as shown in \Fig{fig:hardware}), the position error tolerance for flight control is constrained to approximately $\pm$ 5 cm along each axis at the moment of grasping. This poses a significant challenge for our system, especially during high-speed flight. Nevertheless, our system achieves an 84\% SR across 25 trials, with grasping speeds reaching up to 1.5~m/s. The grasping speed here is defined as the average flight speed at the moment of grasping.
% within a 10~cm distance of the grasping point. 
\Fig{fig:real_curve} illustrates the state evolution of our system during a real-world trial, demonstrating that the policy is able to closely track both the desired position and yaw angle, determine the appropriate timing to close the gripper in advance, and perform grasping in high-speed flight. These results highlight the robustness and agility of our system, on par with state-of-the-art aerial grasping systems based on traditional pipelines and sophisticated grippers. Demonstrations of the real-world experiments can be found in the supplementary video.
% The real-world experiments are also shown in the supplementary video.

\begin{figure}[h]
\centering
\includegraphics[width=0.45\textwidth,trim={0.5cm 6.5cm 0.5cm 7cm},clip]{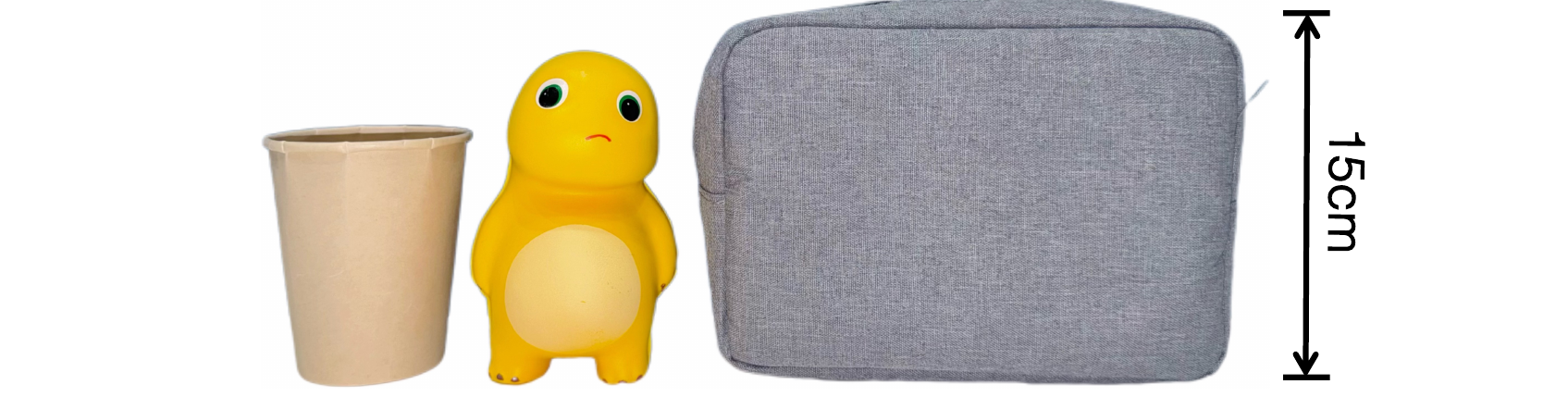} % trim order: left bottom right top
\caption{The three objects we use for additional grasping experiments. From left to right: cup, rubber toy and pouch.}

\label{fig:object}
\end{figure}

Furthermore, we evaluated our platform on grasping different objects, including a small cup, a rubber toy, and a pouch. \Fig{fig:object} illustrates their geometries, while Table~\ref{tab:ObjSuccessRates} summarizes their masses and dimensions. 
These objects, approximately 7~cm wide, require precise position control given the gripper’s 12~cm maximum opening width, which allows only a positional tolerance of approximately $\pm$ 2.5~cm at the moment of grasping. However, our system achieved a grasp success rate of 9/10, 9/10, and 8/10 for the cup, rubber toy, and pouch, respectively. These results demonstrate that our approach can generalize to grasping objects with different shapes and masses, and further verify the control accuracy and robustness of our system.
% During experiments, we also observed that when grasping objects with large surface areas, such as the pouch, the quadrotor exhibited a noticeable drop due to the object occluding a large portion of the thrust envelope.

% \vspace{-0.3cm}

Due to the effectiveness of the OTE module, sim-to-real transfer techniques such as domain randomization and sensor noise were not employed, as these methods did not lead to obvious performance improvement and instead complicated the policy training.
The failure cases in the tests were primarily due to imprecise flight control, causing the gripper to collide with the object. We attribute this to the difference in quadrotor dynamics models between the Crazyflie 2.x used in simulation training \cite{panerati_learning_2021} and the customized real-world platform. More refined system identification and modeling could further narrow the sim-to-real gap, thereby improving grasping performance.

\begin{table}[h]
\scriptsize
    \renewcommand{\arraystretch}{1.3}
    \caption{Grasping results of different objects}
    \centering
    \begin{tabular}{@{\extracolsep{\fill}} c c c c @{}}
        \toprule
        \makecell{Object} & \makecell{Success Rate} & \makecell{Weight [g]} & \makecell{Dimensions [cm]} \\
        \midrule
        Cup & 9/10 & 5 & 7.5$\times$7.5$\times$8.5  \\
        Rubber Toy & 9/10 & 41 & 8.0$\times$7.0$\times$12.0 \\
        Pouch & 8/10 & 90 & 19.0$\times$6.5$\times$15.0 \\
        \bottomrule
    \end{tabular}
    \label{tab:ObjSuccessRates}
\end{table}

\vspace{-0.2cm}
\section{Conclusion and Future Work} \label{sec:conclusions}
In this work, we present Swooper, a DRL-based approach for high-speed aerial grasping. Swooper adopts a two-stage learning strategy that first trains a flight control policy and then fine-tunes it to acquire grasping skills. 
The policy trained in simulation demonstrates zero-shot sim-to-real transfer on our customized platform.

Our method unifies precise flight control and active manipulation into a single lightweight policy. Optimization-based approaches such as Model Predictive Control (MPC) can also potentially handle both flight and gripper control simultaneously. However, jointly solving these control objectives often imposes high computational costs, which may limit real-time performance. On the other hand, the potential to be extended to vision-based aerial grasping represents one of the unique strengths of our DRL-based framework.

Our work demonstrates the feasibility of achieving high-speed aerial grasping using DRL-based approaches. 
% Future work could focus on developing an onboard system for self-localization and target object localization to eliminate reliance on the Vicon system, enabling both indoor and outdoor grasping similar to \cite{bauer_open-source_2024}. 
Future work could focus on more refined quadrotor system identification and dynamics modeling, which would help bridge the sim-to-real gap and enhance aerial grasping performance. Another path for future research is to address ground effects and enhance the system’s resistance to external disturbances during  grasping using DRL-based approaches. Finally, vision-based, end-to-end aerial grasping remains a challenging yet worthwhile avenue for further investigation.

\bibliographystyle{IEEEtran}
\bibliography{refers}

\begin{thebibliography}{10}
\providecommand{\url}[1]{#1}
\csname url@rmstyle\endcsname
\providecommand{\newblock}{\relax}
\providecommand{\bibinfo}[2]{#2}
\providecommand\BIBentrySTDinterwordspacing{\spaceskip=0pt\relax}
\providecommand\BIBentryALTinterwordstretchfactor{4}
\providecommand\BIBentryALTinterwordspacing{\spaceskip=\fontdimen2\font plus
\BIBentryALTinterwordstretchfactor\fontdimen3\font minus \fontdimen4\font\relax}
\providecommand\BIBforeignlanguage[2]{{%
\expandafter\ifx\csname l@#1\endcsname\relax
\typeout{** WARNING: IEEEtran.bst: No hyphenation pattern has been}%
\typeout{** loaded for the language `#1'. Using the pattern for}%
\typeout{** the default language instead.}%
\else
\language=\csname l@#1\endcsname
\fi
#2}}

\bibitem{bauer_open-source_2024}
E.~Bauer, M.~Bl{\"o}chlinger, P.~Strauch, A.~Raayatsanati, C.~Curdin, and R.~K. Katzschmann, ``An open-source soft robotic platform for autonomous aerial manipulation in the wild,'' in \emph{Proc. Conf. Robot. Learn.}, 2024.

\bibitem{aucone_embodied_2025}
E.~Aucone and S.~Mintchev, ``Embodied aerial physical interaction: Combining body and brain for robust interaction with unstructured environments,'' \emph{Sci. Robot.}, vol.~10, no. 102, p. eads0200, 2025.

\bibitem{song_reaching_2023}
Y.~Song, A.~Romero, M.~M{\"u}ller, V.~Koltun, and D.~Scaramuzza, ``Reaching the limit in autonomous racing: Optimal control versus reinforcement learning,'' \emph{Sci. Robot.}, vol.~8, no.~82, p. eadg1462, 2023.

\bibitem{song_autonomous_2021}
Y.~Song, M.~Steinweg, E.~Kaufmann, and D.~Scaramuzza, ``Autonomous drone racing with deep reinforcement learning,'' in \emph{Proc. IEEE Int. Conf. Intell. Robots Syst.}, 2021, pp. 1205--1212.

\bibitem{penicka_learning_2022}
R.~Penicka, Y.~Song, E.~Kaufmann, and D.~Scaramuzza, ``Learning minimum-time flight in cluttered environments,'' \emph{IEEE Robot. Automat. Lett.}, vol.~7, no.~3, pp. 7209--7216, 2022.

\bibitem{kaufmann_champion-level_2023}
E.~Kaufmann, L.~Bauersfeld, A.~Loquercio, M.~M{\"u}ller, V.~Koltun, and D.~Scaramuzza, ``Champion-level drone racing using deep reinforcement learning,'' \emph{Nature}, vol. 620, no. 7976, pp. 982--987, 2023.

\bibitem{xing2024bootstrapping}
J.~Xing, A.~Romero, L.~Bauersfeld, and D.~Scaramuzza, ``Bootstrapping reinforcement learning with imitation for vision-based agile flight,'' in \emph{Proc. Conf. Robot. Learn.}, 2024.

\bibitem{xing2024multi}
J.~Xing, I.~Geles, Y.~Song, E.~Aljalbout, and D.~Scaramuzza, ``Multi-task reinforcement learning for quadrotors,'' \emph{IEEE Robot. Automat. Lett.}, 2024.

\bibitem{wang_agile_2024}
M.~Wang, S.~Jia, Y.~Niu, Y.~Liu, C.~Yan, and C.~Wang, ``Agile flights through a moving narrow gap for quadrotors using adaptive curriculum learning,'' \emph{IEEE Trans. Intell. Veh.}, 2024.

\bibitem{wu_whole-body_2024}
T.~Wu, Y.~Chen, T.~Chen, G.~Zhao, and F.~Gao, ``Whole-body control through narrow gaps from pixels to action,'' in \emph{Proc. IEEE Int. Conf. Robot. Autom.}, 2025, pp. 11\,317--11\,324.

\bibitem{kooi_inclined_2021}
J.~E. Kooi and R.~Babu{\v{s}}ka, ``Inclined quadrotor landing using deep reinforcement learning,'' in \emph{Proc. IEEE Int. Conf. Intell. Robots Syst.}, 2021, pp. 2361--2368.

\bibitem{ladosz_autonomous_2024}
P.~Ladosz, M.~Mammadov, H.~Shin, W.~Shin, and H.~Oh, ``Autonomous landing on a moving platform using vision-based deep reinforcement learning,'' \emph{IEEE Robot. Automat. Lett.}, vol.~9, no.~5, pp. 4575--4582, 2024.

\bibitem{huang_earl_2023}
B.~Huang, J.~Yu, and S.~Jain, ``Earl: Eye-on-hand reinforcement learner for dynamic grasping with active pose estimation,'' in \emph{Proc. IEEE Int. Conf. Intell. Robots Syst.}, 2023, pp. 2963--2970.

\bibitem{hu_grasping_2023}
Z.~Hu, Y.~Zheng, and J.~Pan, ``Grasping living objects with adversarial behaviors using inverse reinforcement learning,'' \emph{IEEE Trans. Robot.}, vol.~39, no.~2, pp. 1151--1163, 2023.

\bibitem{orsula_learning_2022}
A.~Orsula, S.~B{\o}gh, M.~Olivares-Mendez, and C.~Martinez, ``Learning to grasp on the moon from 3d octree observations with deep reinforcement learning,'' in \emph{Proc. IEEE Int. Conf. Intell. Robots Syst.}, 2022, pp. 4112--4119.

\bibitem{fishman_dynamic_2021}
J.~Fishman, S.~Ubellacker, N.~Hughes, and L.~Carlone, ``Dynamic grasping with a ``soft" drone: From theory to practice,'' in \emph{Proc. IEEE Int. Conf. Intell. Robots Syst.}, 2021, pp. 4214--4221.

\bibitem{Ubellacker2024}
S.~Ubellacker, A.~Ray, J.~M. Bern, J.~Strader, and L.~Carlone, ``High-speed aerial grasping using a soft drone with onboard perception,'' \emph{npj Robot.}, vol.~2, no.~1, p.~5, 2024.

\bibitem{appius_raptor_2022}
A.~X. Appius, E.~Bauer, M.~Bl{\"o}chlinger, A.~Kalra, R.~Oberson, A.~Raayatsanati, P.~Strauch, S.~Suresh, M.~von Salis, and R.~K. Katzschmann, ``Raptor: Rapid aerial pickup and transport of objects by robots,'' in \emph{Proc. IEEE Int. Conf. Intell. Robots Syst.}, 2022, pp. 349--355.

\bibitem{crooks2016fin}
W.~Crooks, G.~Vukasin, M.~O’Sullivan, W.~Messner, and C.~Rogers, ``Fin ray{\textregistered} effect inspired soft robotic gripper: From the robosoft grand challenge toward optimization,'' \emph{Front. Robot. AI}, vol.~3, p.~70, 2016.

\bibitem{dimmig_nonprehensile_2024}
C.~A. Dimmig and M.~Kobilarov, ``Non-prehensile aerial manipulation using model-based deep reinforcement learning,'' in \emph{IEEE Int. Conf. Autom. Sci. Eng.}, 2024, pp. 2194--2200.

\bibitem{Kaufmann_benchmark_2022}
E.~Kaufmann, L.~Bauersfeld, and D.~Scaramuzza, ``A benchmark comparison of learned control policies for agile quadrotor flight,'' in \emph{Proc. IEEE Int. Conf. Robot. Autom.}, 2022, pp. 10\,504--10\,510.

\bibitem{shahid_learning_2020}
A.~A. Shahid, L.~Roveda, D.~Piga, and F.~Braghin, ``Learning continuous control actions for robotic grasping with reinforcement learning,'' in \emph{Proc. IEEE Int. Conf. Syst. Man Cybern.}, 2020, pp. 4066--4072.

\bibitem{schulman2017proximal}
J.~Schulman, F.~Wolski, P.~Dhariwal, A.~Radford, and O.~Klimov, ``Proximal policy optimization algorithms,'' \emph{arXiv:1707.06347}, 2017.

\bibitem{panerati_learning_2021}
J.~Panerati, H.~Zheng, S.~Zhou, J.~Xu, A.~Prorok, and A.~P. Schoellig, ``Learning to fly—a gym environment with pybullet physics for reinforcement learning of multi-agent quadcopter control,'' in \emph{Proc. IEEE Int. Conf. Intell. Robots Syst.}, 2021, pp. 7512--7519.

\bibitem{coumanspybullet}
E.~Coumans and Y.~Bai, ``Pybullet, a python module for physics simulation for games, robotics and machine learning,'' 2016.

\bibitem{stable-baselines3}
A.~Raffin, A.~Hill, A.~Gleave, A.~Kanervisto, M.~Ernestus, and N.~Dormann, ``Stable-baselines3: Reliable reinforcement learning implementations,'' \emph{J. Mach. Learn. Res.}, vol.~22, no. 268, pp. 1--8, 2021.

\bibitem{liu2024omninxt}
P.~Liu, C.~Feng, Y.~Xu, Y.~Ning, H.~Xu, and S.~Shen, ``Omninxt: A fully open-source and compact aerial robot with omnidirectional visual perception,'' in \emph{Proc. IEEE Int. Conf. Intell. Robots Syst.}, 2024, pp. 10\,605--10\,612.

\bibitem{zhang_back_2024}
Y.~Zhang, Y.~Hu, Y.~Song, D.~Zou, and W.~Lin, ``Learning vision-based agile flight via differentiable physics,'' \emph{Nat. Mach. Intell.}, pp. 1--13, 2025.

\bibitem{thomas_avian-inspired_2013}
J.~Thomas, J.~Polin, K.~Sreenath, and V.~Kumar, ``Avian-inspired grasping for quadrotor micro uavs,'' in \emph{Proc. ASME Int. Des. Eng. Tech. Conf. Comput. Inf. Eng. Conf}, vol. 55935, 2013, p. V06AT07A014.

\end{thebibliography}

\end{document}